\documentclass[letterpaper]{article} 
\usepackage{aaai24}  
\usepackage{times}  
\usepackage{helvet}  
\usepackage{courier}  
\usepackage[hyphens]{url}  
\usepackage{graphicx} 
\usepackage{natbib}  
\usepackage{caption} 
\usepackage{algorithm}
\usepackage{algorithmic}

\usepackage{cite}
\usepackage{amsmath,amssymb,amsfonts}
\usepackage{graphicx}
\usepackage{textcomp}
\usepackage{xcolor}
\usepackage{tabularx}
\usepackage{subcaption}
\usepackage{booktabs}
\usepackage{multirow}
\usepackage{arydshln}

\definecolor{nred}{RGB}{196, 38, 11}
\definecolor{nblue}{RGB}{41, 52, 190}
\definecolor{ngreen}{RGB}{18, 141, 21}

\usepackage{newfloat}
\usepackage{listings}
\DeclareCaptionStyle{ruled}{labelfont=normalfont,labelsep=colon,strut=off} 
\floatstyle{ruled}
\newfloat{listing}{tb}{lst}{}
\floatname{listing}{Listing}
\begin{document}


\title{On the Cultural Gap in Text-to-Image Generation}
\author {
    Bingshuai Liu,\textsuperscript{\rm 1,2}\thanks{Equal contribution: Bingshuai Liu and Longyue Wang. Work done while Bingshuai Liu and Chengyang Lyu were interning at Tencent AI Lab.}
    Longyue Wang,\textsuperscript{\rm 1}$^*$
    Chenyang Lyu,\textsuperscript{\rm 1,3}
    Yong Zhang\textsuperscript{\rm 1}\\
    Jinsong Su,\textsuperscript{\rm 2}
    Shuming Shi,\textsuperscript{\rm 1}
    Zhaopeng Tu\textsuperscript{\rm 1}\thanks{Zhaopeng Tu is the corresponding author.}
}
\affiliations {
    \textsuperscript{\rm 1} Tencent AI Lab, \quad
    \textsuperscript{\rm 2} Xiamen University, \quad
    \textsuperscript{\rm 3} Dublin City University\\
    \textsuperscript{\rm 1}\{vinnylywang,norriszhang,shumingshi,zptu\}@tencent.com,  \quad \textsuperscript{\rm 2}\{bsliu,jssu\}@ximen.com, \quad \textsuperscript{\rm 3}cylyu@dcu.ie
}

\maketitle

\begin{abstract}

One challenge in text-to-image (T2I) generation is the inadvertent reflection of culture gaps present in the training data, which signifies the disparity in generated image quality when the cultural elements of the input text are rarely collected in the training set. Although various T2I models have shown impressive but arbitrary examples, there is no benchmark to systematically evaluate a T2I model's ability to generate cross-cultural images. To bridge the gap, we propose a Challenging Cross-Cultural (C$^3$) benchmark with comprehensive evaluation criteria, which can assess how well-suited a model is to a target culture. By analyzing the flawed images generated by the Stable Diffusion model on the C$^3$ benchmark, we find that the model often fails to generate certain cultural objects. Accordingly, we propose a novel multi-modal metric that considers object-text alignment to filter the fine-tuning data in the target culture, which is used to fine-tune a T2I model to improve cross-cultural generation. Experimental results show that our multi-modal metric provides stronger data selection performance on the C$^3$ benchmark than existing metrics, in which the object-text alignment is crucial. 
We release the benchmark, data, code, and generated images to facilitate future research on culturally diverse T2I generation.\footnote{\url{https://github.com/longyuewangdcu/C3-Bench}.}

\end{abstract}

\section{Introduction}

\begin{figure}[t]
  \centering
  \begin{subfigure}[b]{0.225\textwidth}
    \includegraphics[width=\textwidth]{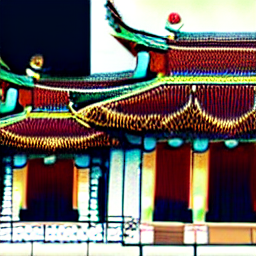}
    \label{fig:example_subfig_a}
  \end{subfigure}
  \hfill
  \begin{subfigure}[b]{0.225\textwidth}
    \includegraphics[width=\textwidth]{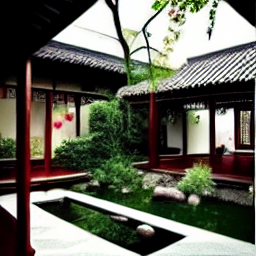}
    \label{fig:example_subfig_b}
  \end{subfigure}
  \caption{Comparison of the original stable diffusion (left) and the stable diffusion fine-tuned on the dataset filtered by our approach (right) for generating cross-cultural images with Chinese elements based on the prompt \textit{A garden with typical Chinese architecture and design elements}. The example clearly demonstrates that the fine-tuned system can produce higher quality images.}
  \label{fig:example}
\end{figure}

Text-to-image (T2I) generation has emerged as a significant research area in recent years, with numerous applications spanning advertising, content creation, accessibility tools, human-computer interaction, language learning, and cross-cultural communication~\cite{rombach2022high}. 
One challenge of T2I models is the inadvertent reflection or amplification of cultural gaps present in the training data, which refer to differences in norms, values, beliefs, and practices across various cultures~\cite{prabhakaran2022cultural,struppek2022biased}. 
The cultural gap in T2I generation signifies the disparity in image generation quality when the cultural elements of the input text are rarely collected in the training set. For example, in the LAION 400M dataset, the collected text-image pairs predominantly consist of English texts and images containing Western cultural elements. Consequently, when given a text description featuring Eastern cultural elements, the quality of the generated image is likely to be unsatisfactory.
Figure~\ref{fig:example} shows an example. The Stable Diffusion v1-4 model that is trained on the Western cultural data fails to generate satisfying Chinese cultural elements.

The lack of cultural sensitivity in the generated images can manifest in the form of images that may be inappropriate, offensive, or simply irrelevant in certain cultural contexts. Therefore, addressing these cultural gaps in AI T2I models is crucial to ensure the generation of culturally appropriate and contextually relevant images for users from diverse cultural backgrounds.
However, although various T2I models have shown how the cultural gap leads to flawed images with impressive but arbitrary examples, there is no benchmark to systematically evaluate a T2I model's ability to generate cross-cultural images.

To bridge the gap, we introduce a C$^3$ benchmark with comprehensive evaluation criteria for the target evaluation on the cross-cultural T2I generation. 
Given that current open-sourced T2I models are generally trained on the English data associated with Western cultural elements~\cite{rombach2022high,ramesh2022hierarchical}, we built a evaluation set of textual prompts designed for generating images in Chinese cultural style. Specifically, we ask the powerful GPT-4 model with carefully designed context to generate the challenging prompts that can lead a T2I model to make different types of mistakes for cross-cultural generation. We also provide a set of evaluation criteria that consider characteristics (e.g. cultural appropriateness) and challenges (e.g. cross-cultural object presence and localization) of cross-cultural T2I generation.

A promising way of improving cross-cultural generation is to fine-tune a T2I model on training data in target culture, which are generally in other non-English languages. Accordingly, the captions in the target-cultural data are translated to English with external translation systems, which may introduce translation mistakes that can affect the quality of the image-caption pairs. In response to this problem, we propose a novel multi-modal metric that considers both textual and visual elements to filter low-quality translated captions. 
In addition, analyses of generated images on the C$^3$ benchmark show that the object generation in target culture is one of the key challenges for cross-culture T2I generation. Accordingly, our multi-modal metric includes an explicit object-text alignment score to encourage that all necessary objects in the image are included in the translated caption.
Empirical analysis shows that our metric correlates better with human judgement on assessing the quality of translated caption for T2I than existing metrics. 
Experimental results on the C$^3$ benchmark show that our multi-modal metric provides stronger data selection performance.


In summary, our contributions are as follows:
\begin{itemize}
\item We build a benchmark with comprehensive evaluation criteria for cross-cultural T2I generation, which is more challenging than the commonly-used MS-COCO benchmark with more cross-cultural objects.

\item We propose a multi-modal metric that considers both textual and visual elements to filter training data in the target culture, which produce better performance for fine-tuning a T2I model for cross-cultural generation.

\item To facilitate future research on culturally diverse T2I generation, we publicly release the resources we constructed in this paper, including the C$^3$ benchmark, translated dataset, the filtering scripts, and generated images.

\end{itemize}

\section{Related Work} 

In the last several years, there has been a growing interest in T2I generation.
The conventional generation models are built upon generative adversarial networks (GANs)~\cite{reed2016generative, xu2018attngan, zhang2017stackgan}, which consists of a text encoder and an image generator. 
Recently, diffusion models have advanced state of the art in this field by improving image quality and diversity~\cite{ramesh2022hierarchical, ramesh2021zero, rombach2022high, saharia2022photorealistic}. Previous research on text-guided image generation mainly focused on improving the understanding of complex text descriptions~\cite{zhu2019dm,ruan2021dae} or the quality of generated images~\cite{saharia2022photorealistic}. 
In this work, we aim to improve the generalization of T2I models to generate images associated with cultural elements that have rarely been observed in the training data.

Another thread of research turns to enhance multilingual capabilities of T2I models, which can support non-English input captions. For example, 
~\citet{chen2022altclip} extent the text encoder of diffusion model with a pre-trained multilingual text encoder XLM-R.
\citet{li2023translation} mitigated the language gap by translating English captions to other languages with neural machine translation systems.
While they enhance the multilingual support for the textual caption, we focus on improving the ``multilingual support'' for the generated images by enabling the model to generate images of cultural content in different languages. 

\citet{saxon2023multilingual} proposed a novel approach for benchmarking the multilingual parity of generative T2I systems by assessing the ``conceptual coverage'' of a model across different languages. They build an atomic benchmark that narrowly and reliably captures a specific characteristic -- conceptual knowledge as reflected by a model's ability to reliably generate images of an object across languages. Similarly, we build a benchmark to capture another specific characteristic -- cross-cultural generation as reflected by a model's ability to reliably generate cultural elements that are rarely collected in the training set.

Closely related to this work, \citet{chen2022pali} introduced the PaLI model, which is trained on a large multilingual mix of pre-training tasks containing 10B images and texts in over 100 languages. This model emphasizes the importance of scale in both the visual and language parts of the model and the interplay between the two. Our work is complementary to theirs: we provide a benchmark for the target evaluation on the cross-cultural T2I generation, which can estimate how well-suited a model is to a target cultural. In addition, we propose a novel multi-modal alignment approach for fast adaption of a generation model to a target culture.

\section{Cross-Cultural Challenging (C$^3$) Benchmark}

\subsection{Constructing the C$^3$ Benchmark with GPT-4}

To generate captions for creating cross-cultural and culturally diverse images, we first asked GPT-4 to identify the types of mistakes T2I generation systems can make if they are asked to generate such cross-cultural images. We received the following insights from GPT-4, which serve as the prompt for GPT-4 to generate more challenging captions:
\begin{itemize}
    \item {\em Language Bias}: T2I systems that do not account for variations in regional dialects or Chinese script may generate text that is linguistically inaccurate or insensitive to Chinese language subtleties.
    \item {\em Cultural Inappropriateness}: Without an accurate understanding of Chinese cultural norms and values, a T2I generation system may generate images that are seen as inappropriate or offensive.
    \item {\em Missed Cultural Nuances}: T2I systems that lack an appreciation for the nuances of Chinese culture may generate images that are not authentic or credible.
    \item {\em Stereotyping and Counterfeit Representations}: T2I systems that rely on popular stereotypes or inaccurate depictions of Chinese culture may generate images that perpetuate damaging myths, or counterfeit representations give mistaken impressions.
    \item {\em Insufficient Diversity}: A T2I system that does not consider the diversity of China's 56  ethnic groups or pay attention to minority cultures' rich heritage may overgeneralize or oversimplify Chinese culture.
\end{itemize}

\begin{table}[t]
\centering
\caption{Five seed captions for constructing benchmark.}
\begin{tabular}{m{7.8cm}}
\hline
A family enjoying a feast of traditional American fast food while sitting on a Chinese-style bamboo mat \\
\hline
A group of people performing a dragon dance at the opening of a new European-style cafe \\
\hline
A portrait of a woman wearing a beautiful qipao dress, holding a plate of hamburgers and fries \\
\hline
A bustling scene at a village fair, showcasing both Chinese lanterns and Western-style carnival games \\
\hline
An ancient Chinese temple adorned with modern neon signs advertising various global brands \\
\hline
\end{tabular}
\label{tab:seed}
\end{table}

Subsequently, we asked GPT-4 to provide five representative examples of image captions in English that could lead a T2I system, trained only on English data, to make different types of mistakes when generating images reflecting Chinese culture or elements, as listed in Table~\ref{tab:seed}. 
We used the first five examples (selected and checked by humans) as seed examples to iteratively generate more diverse and different examples, which can lead to errors while generating images reflecting Chinese culture or elements.
Specifically, we use the following prompt to obtain more challenging captions:

\vspace{5pt}
\noindent\fbox{\begin{minipage}{0.96\linewidth}

T2I systems trained only on English data can make mistakes when generating images reflecting Chinese culture/element:

Language bias: T2I systems that do not account $\cdots$

$\cdots$ 

may overgeneralize or oversimplify Chinese culture.

\vspace{5pt}
{\em Can you give five representative image captions in English that could lead a T2I generation trained only on English data make different types of mistakes above when generating images reflecting Chinese culture/element based on the examples but different from the examples below: {}}

\vspace{5pt}
Please follow the format and only give me captions (the captions do not have to contain the word `Chinese'), no other texts:

Example 1: Caption1

$\cdots$

Example 5: Caption5
\end{minipage}}

\begin{figure}[t]
  \centering
  \subfloat[\bf Data Statistics]
  {\begin{tabular}{c rrr}
    \toprule
    \bf     &  \bf C$^3$   &  \bf C$^3$+    &   \bf COCO\\
    \hline
    \bf Caption &   500   &   9,889     &   500\\
    \bf Length   &  29.34 &   26.49     &   10.22 \\
    \bf Object   &  10.76 &   9.81      &   3.65\\
    \bottomrule
    \end{tabular}}
  \\
  \subfloat[\bf C$^3$]
  {\includegraphics[width=0.15\textwidth]{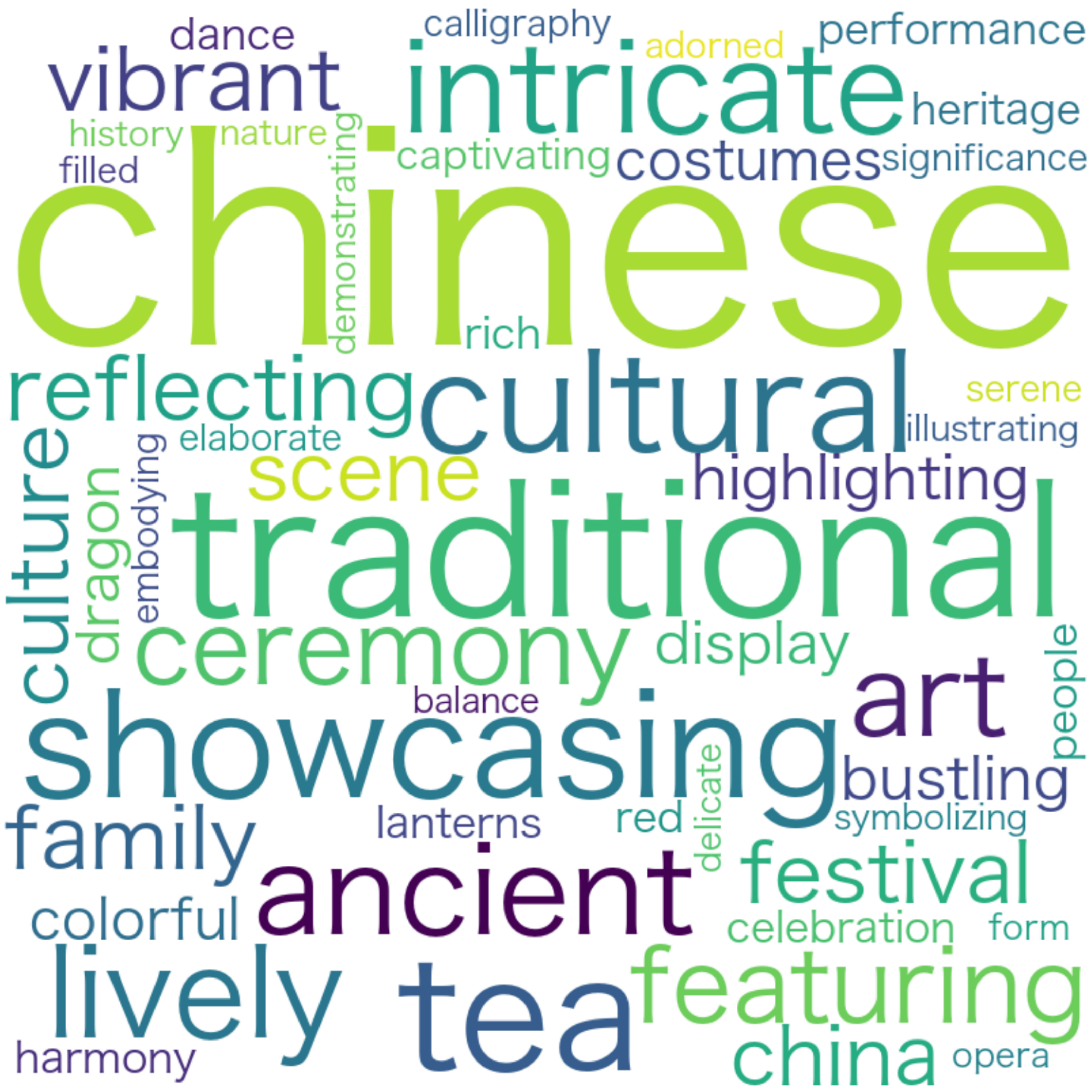}}
  \hfill
  \subfloat[\bf C$^3$+]
  {\includegraphics[width=0.15\textwidth]{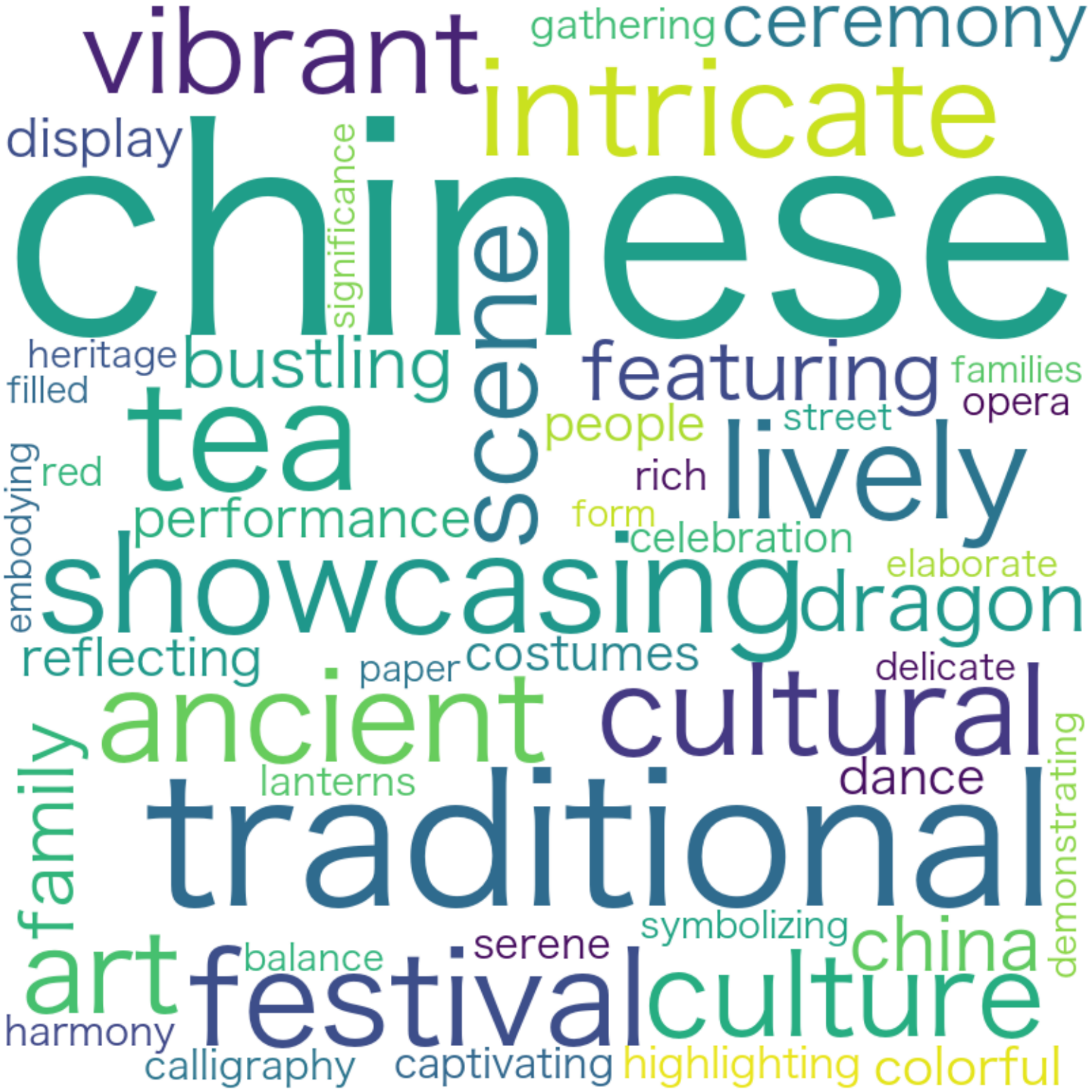}}
  \hfill
  \subfloat[\bf MS-COCO]
  {\includegraphics[width=0.15\textwidth]{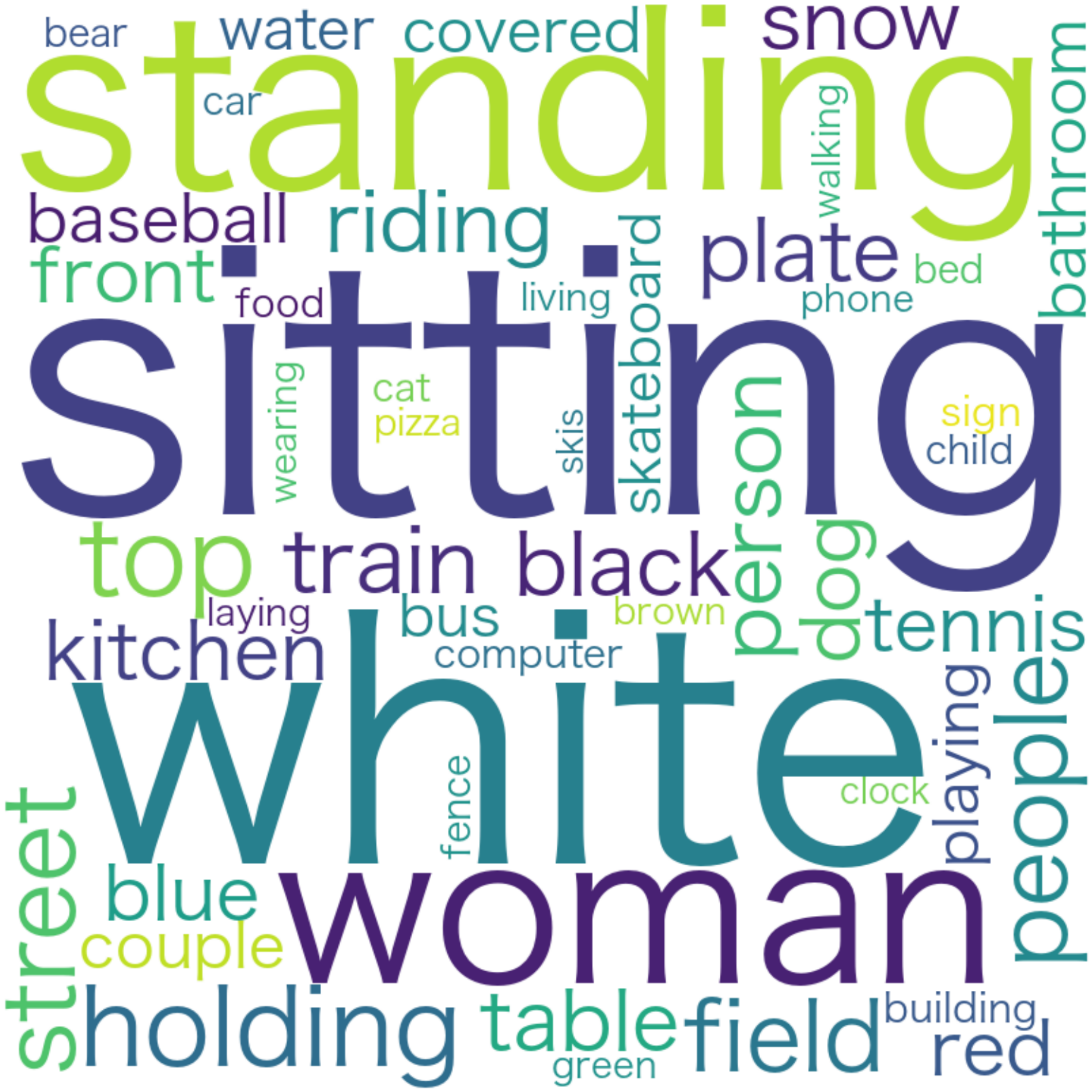}}
  \caption{Statistics (a) and Word Cloud (b,c) of the C$^3$ benchmark and its expanded edition C$^3$+. ``Length'' and ``Object'' denote the average number of words and objects in each caption, respectively. We list the details of the MS-COCO Captions (``MS-COCO'') benchmark for reference.}
  \label{fig:benchmark_details}
\end{figure}

\vspace{5pt}
In each iteration we randomly sample five seed examples from the generated examples as prompt examples. The collected image captions were used to construct an evaluation set for assessing the performance of T2I generation systems in generating cross-cultural and culturally diverse images.
Finally, we obtain a set of $9,889$ challenging captions by filtering the repetitive ones for cross-cultural T2I generation, which we name as C$^3$+. 
Since it is time-consuming and labor-intensive to manually evaluate the generated images for all the captions, we randomly sample 500 captions to form a small-scale benchmark C$^3$, which will serve as the testbed in the following experiments for human evaluation. The generated images for different models on the full C$^3$+ benchmark (without human evaluation) will also be released for future research.
Figure~\ref{fig:benchmark_details} shows the benchmark details.

\subsection{Evaluating Difficulty of the C$^3$ Benchmark}

\begin{figure}[t]
  \centering
    \includegraphics[width=0.45\textwidth]{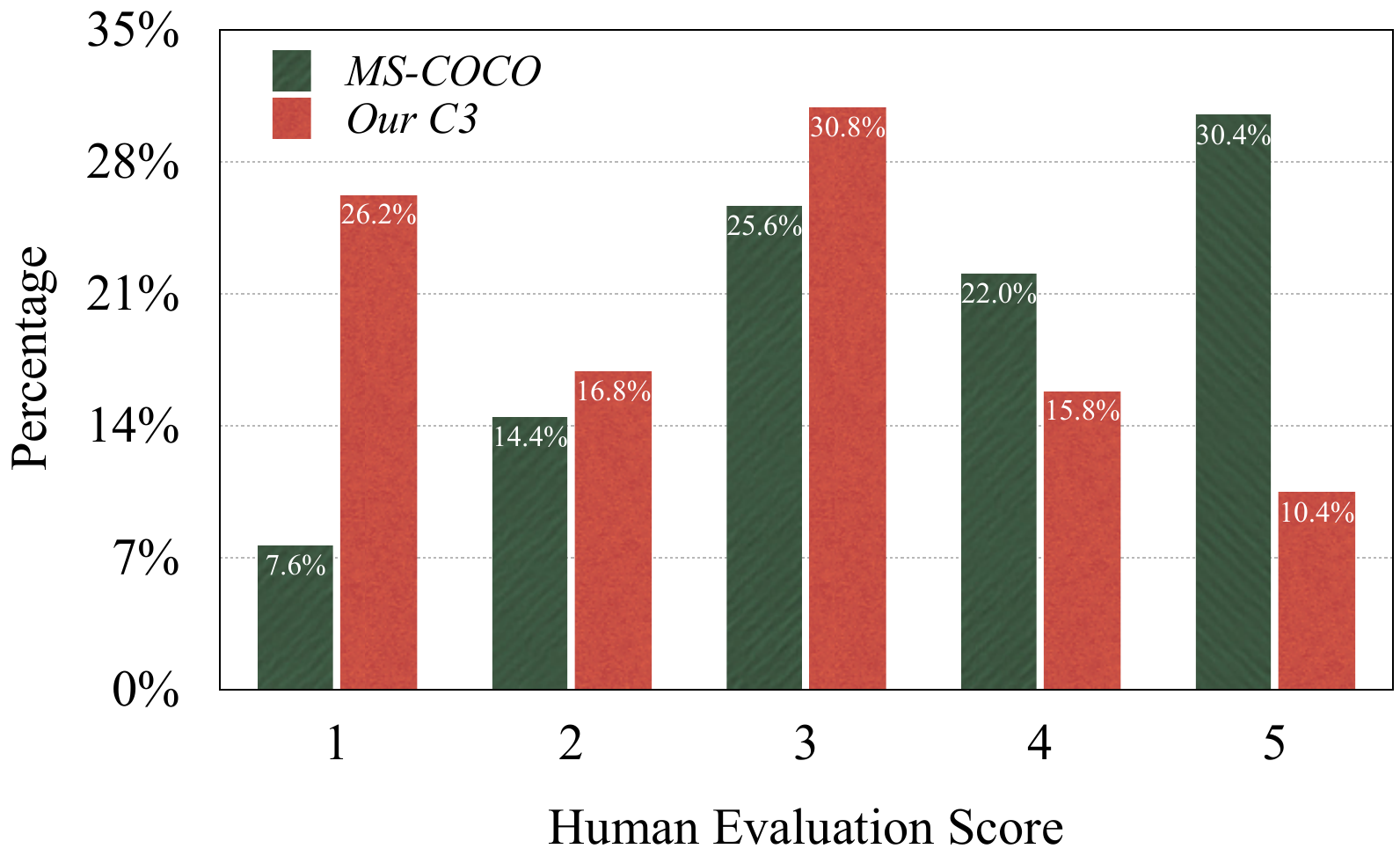}
  \caption{Human scoring results of Stable Diffusion on the widely-used MS-COCO and the proposed C$^3$ benchmarks. 
  }
  \label{fig:compare_coco_benchmark}
\end{figure}

\begin{figure}[t]
\centering
\subfloat[\bf MS-COCO Benchmark]{
\begin{tabular}{m{0.2\textwidth}m{0.2\textwidth}}
\includegraphics[width=0.2\textwidth]{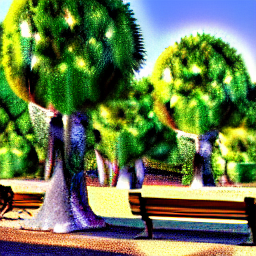}  &
\includegraphics[width=0.2\textwidth]{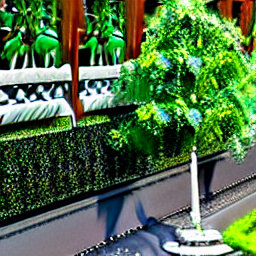}\\
\multicolumn{2}{m{0.45\textwidth}}{(1) A park bench in the midst of a beautiful desert garden.}\\
\multicolumn{2}{m{0.45\textwidth}}{(2) An outdoor garden area with verdant plants and a tree.}
\end{tabular}
}
\\
\subfloat[\bf C$^3$ Benchmark]{
\begin{tabular}{m{0.2\textwidth}m{0.2\textwidth}}
\includegraphics[width=0.2\textwidth]{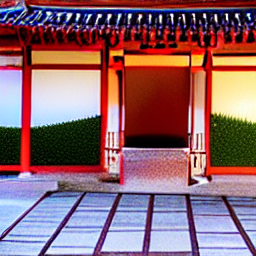}  &
\includegraphics[width=0.2\textwidth]{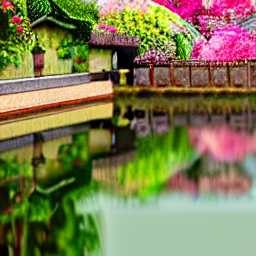}\\
\multicolumn{2}{m{0.45\textwidth}}{(1) A serene scene of {\color{red} a tea ceremony} in a serene Chinese garden setting.} \\
\multicolumn{2}{m{0.45\textwidth}}{(2) A beautiful Chinese garden with {\color{red} a gracefully arched bridge} and {\color{red} blooming lotus flowers}.}\\
\end{tabular}
}
\caption{Example images generated by the Stable Diffusion v1-4 model on the MS-COCO and C$^3$ benchmarks. We highlight in {\color{red} red} the objects missed in the generated image.}
\label{fig:example_diffusion}
\end{figure}

To evaluate the difficulty of the C$^3$ benchmark, we compare with the commonly-used COCO Captions dataset~\cite{chen2015microsoft}, which is extracted from the English data that is potentially similar in distribution with the training data of Stable Diffusion. 
Specifically, we sample 500 captions from the COCO data, and ask the Stable Diffusion v1.4 model to generate images based on the captions.
Figure~\ref{fig:benchmark_details} shows the details of the sampled COCO Caption data. Compared with C$^3$, the captions in COCO contain smaller sizes of words and objects, which makes it easier for T2I generation.

For comparing the quality of the generated images on both benchmarks, we follow the common practices to ask human annotators to score the generated images from the perspectives of both the image-text alignment and image fidelity~\cite{saharia2022photorealistic,feng2023ernie}. 
Figure~\ref{fig:compare_coco_benchmark} lists the comparison results. Clearly, 78\% of the generated images on COCO are rated above average (``$\geq3$''), while the ratio on C$^3$ is 57\%. Specifically, 26.2\% of the generated images on C$^3$ is rated as the lowest 1 score, which is far larger than that on COCO.
Figure~\ref{fig:example_diffusion} shows some examples of generated images on the two benchmarks. The Stable Diffusion model successfully generates all objects in the MS-COCO captions. However, it fails to generate cultural objects (e.g. ``a tea ceremony'', ``{a gracefully arched bridge}'', and ``blooming lotus flowers'') in the C$^3$ captions, which are rarely observed in the training data of the diffusion model.
These results demonstrate that the proposed C$^3$ is more challenging.

\begin{figure*}[t]
  \centering
    \includegraphics[width=0.75\textwidth]{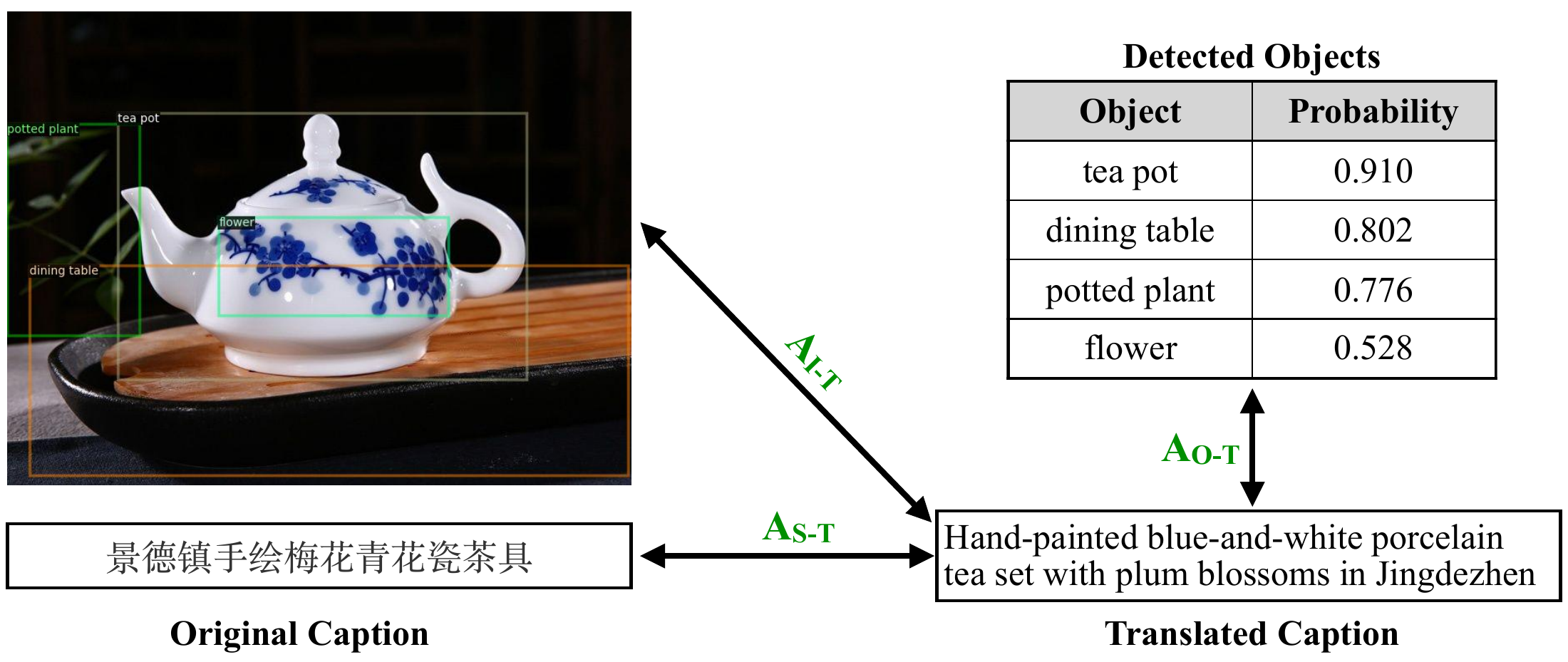}
  \caption{Framework of our filtering metric that measures the quality of the translated caption with three alignment scores: 1) $A_{S-T}$ for aligning the original caption; 2) $A_{I-T}$ for aligning the image; and 3) $A_{O-T}$ for aligning the detected objects.}
  \label{fig:framework}
\end{figure*}

\subsection{Human Evaluation Criteria for the C$^3$ Benchmark}

\begin{table}[t]
\centering
\caption{Evaluation scores for the example image generated by the vanilla stable diffusion model in Figure~\ref{fig:example} (left panel).}
\begin{tabular}{c c m{5cm}}
\toprule
{\bf Criteria}   &   \bf $S$  &  \bf Reasons\\
\midrule
\parbox{1.8cm}{\small \bf Cultural \\ Appropriate} & 3    & The specific cultural elements and styles of China can be distinguished in the image, but there are some meaningless parts.\\
\midrule
\parbox{1.8cm}{\small \bf Object\\ Presence} & 3    & Some objects can be seen in the image, but it is difficult to distinguish specific elements.\\
\midrule
\parbox{1.8cm}{\small \bf Object\\ Localization} & 2    & The temple elements in the image are not lined up correctly.\\
\midrule
\parbox{1.8cm}{\small \bf Semantic\\ Consistency} & 2    & The consistency between the image and the caption is poor.\\
\midrule
\parbox{1.8cm}{\small \bf Visual\\ Aesthetics} & 1    & Overall image quality is very poor.\\
\midrule
\parbox{1.8cm}{\small \bf Cohesion} & 2    & Multiple elements in the image are not coherently matched.\\
\bottomrule
\end{tabular}
\label{tab:example-evaluation}
\end{table}

Although the metrics of image-text alignment and image fidelity are widely-used for general T2I generation, they may not be sufficient to capture the certain types of mistake in the cross-cultural scenario (e.g. cultural inappropriateness and object presence). 
In response to this problem, we propose a fine-grained set of criteria for the target evaluation on the cross-cultural T2I generation, which focuses on various aspects of cultural relevance and image quality:
\begin{enumerate}
\item \textbf{Cultural Appropriateness} that examines the extent to which the generated images reflect the cultural style and context mentioned in the caption. This criterion helps to demonstrate the model's ability to capture and generate culturally relevant visual content.

\item \textbf{Object Presence} that evaluates whether the generated images contain the essential objects mentioned in the caption. This criterion ensures that the model accurately generates the cross-cultural objects in the caption. 

\item \textbf{Object Localization} that assesses the correct placement and spatial arrangement of objects within the generated images, which can be challenging for the cross-cultural objects. This criterion ensures that the model maintains the context and relationships between objects as described in the caption.

\item \textbf{Semantic Consistency} that assesses the consistency between the generated images and the translated captions, ensuring that the visual content aligns with the meaning of the text. This criterion evaluates the model's ability to generate images that accurately represent the caption.

\item \textbf{Visual Aesthetics} that evaluates the overall visual appeal and composition of the generated images. This criterion considers factors such as color harmony, contrast, and image sharpness, which contribute to the perceived quality of the generated images.

\item \textbf{Cohesion} that examines the coherence and unity of the generated images. This criterion evaluates whether all elements appear natural and well-integrated, contributing to a cohesive visual scene.

\end{enumerate}
As seen, in addition to generalizing the conventional image-text alignment (e.g. semantic consistency) and image fidelity (e.g. visual aesthetics and cohesion) criteria, we also propose several novel metrics that consider characteristics (e.g. cultural appropriateness) and challenges (e.g. cross-cultural object presence and localization) of cross-cultural T2I generation. We hope the fine-grained evaluation criteria can provide a comprehensive assessment of the generated images on the proposed C$^3$ benchmark.
Table~\ref{tab:example-evaluation} lists an example of using the criteria to evaluate the image in Figure~\ref{fig:example} (left panel).
Table~\ref{tab:image-evaluate-guideline} in Appendix lists the guideline of using these criteria for human evaluation.

\section{Improving Cross-Cultural Generation}
\label{sec:methodology}

A promising way of improving cross-cultural T2I generation is to fine-tune the diffusion model on the in-domain data (e.g. image-text pairs of Chinese cultural in this work). Generally, the captions of the in-domain data are translated into English, and the pairs of (translated caption, image) are used to fine-tune the diffusion model. The main challenge lies in how to filter low-quality translated captions.

In this section, we first revisit existing filtering methods, which considers only either text-text alignment or image-text alignment. Inspired by recent successes on multi-modal modeling~\cite{lyu2023macaw}, we propose a novel filtering approach that considers {\bf multi-modal alignment} including both text-text and image-text alignment, as well as explicit object-text alignment since the objects are one of the key challenges for cross-cultural T2I generation.

\subsection{Revisiting Existing Methods}

\paragraph{Text-Text Alignment}

Since there is no reference translation for captions of in-domain data, conventional metrics such as BLEU and Meteor that rely on the reference are unsuitable for evaluating the quality of the translated captions. Accordingly, researchers turn to reference-free metric such as BertScore~\cite{zhang2019bertscore}, which computes a similarity score for two sentences in the same language by leveraging the pre-trained contextual embeddings from BERT.
Along this direction,~\citet{feng-etal-2022-language} propose a multilingual version -- LaBSE, which can compute a similarity score for two sentences in different languages.

\paragraph{Image-Text Alignment}
Another thread of research uses multi-modal pre-trained vision-language models to measure the alignment between caption and images. One representative work is CLIP~\cite{radford2021learning}, which computes a similarity score for a sentence and image with a pre-trained model on a dataset of 400 million (image, text) pairs.

\vspace{5pt}
While prior studies use only either text-text alignment or image-text alignment for filtering the in-domain data, they miss the useful information from the other alignment. In response to this problem, we propose a multi-modal alignment approach to better measure the quality of the (image, translated caption) pair.

\subsection{Our Approach -- Multi-Modal Alignment}

As shown in Figure~\ref{fig:framework}, our filtering metric consists of three types of alignment scores:
\begin{itemize}
    \item {\em Text-Text Alignment} $A_{S-T}$ between the original and the translated captions;
    \item {\em Image-Text Alignment} $A_{I-T}$ between the image and the translated caption;
    \item {\em Object-Text Alignment} $A_{O-T}$ between the detected objects in the image and the translated caption.
\end{itemize}

\begin{table*}[h]
\centering
\caption{Pearson correlation~($p<0.01$) with sentence-level human judgments from different perspectives. ``All'' denotes the overall Pearson correlation in all criteria. ``$-A_{O-T}$'' denotes removing the object-text alignment score $A_{O-T}$ from our metric.}
\begin{tabular}{lrrrrrrr}
\toprule
\bf Filtering & \multicolumn{3}{c}{\bf Textual Translation Quality}  &   \multicolumn{3}{c}{\bf Image Correlation} &  \multirow{2}{*}{\textbf{All}} \\
\cmidrule(lr){2-4}\cmidrule(lr){5-7}
\bf Metric  & \textbf{Adequacy} & \textbf{Fluency} & \textbf{Consistency} & \textbf{Relevance} & \textbf{Context}  & \textbf{Appropriateness} \\
\midrule
LaBSE & 0.107 & -0.033 & 0.194 & 0.167 & 0.215 & 0.125 & 0.129 \\
CLIP & -0.081 & -0.114 & -0.092 & -0.085 & -0.057 & -0.086 & -0.086 \\
\midrule
Ours & \bf 0.220 & \bf 0.149 & \bf 0.295 & \bf 0.220 & \bf 0.215 & 0.163 & \bf 0.211 \\
~~~ $-A_{O-T}$ & 0.098 & -0.050 & 0.185 & 0.158 & 0.211 & 0.115 & 0.119 \\
\hdashline
$A_{O-T}$  &  0.210 & 0.161 & 0.274 & 0.200 & 0.186 & 0.148 & 0.197 \\
\bottomrule
\end{tabular}
\label{tab:direct_evaluation_translation}
\end{table*}

Formally, let $S=\{x_1, \cdots, x_M\}$ be the original non-English caption associated with the image $I$, $T=\{y_1, \cdots, y_N\}$ be the translated caption in English, and $O=\{o_1, \cdots, o_K\}$ be the list of the objects (listed in natural language) detected in the image $I$.
We first encode the captions and objects with a multilingual BERT $\mathcal{E} \in \mathbb{R}^h$~\cite{bert} to the corresponding representations:
\begin{align}
    {\bf H}_{S} &= \mathcal{E}(S)  &   \in  \mathbb{R}^{M \times h}\\
    {\bf H}_{T} &= \mathcal{E}(T)  &   \in  \mathbb{R}^{N \times h}\\
    {\bf H}_{O} &= \mathcal{E}(O)  &   \in  \mathbb{R}^{K \times h}
\end{align}
We encode the image $I$ with a Vision Transformer $\mathcal{V} \in \mathbb{R}^h$~\cite{vit} into a representation vector:
\begin{align}
    {\bf h}_{I} &= \mathcal{V}(I)  &   \in  \mathbb{R}^{h}
\end{align}

We follow~\cite{zhang2019bertscore} to calculate the text-text alignment between two captions as a sum of cosine similarities between their tokens' embeddings:
\begin{align}
    A_{S-T} &= \frac{1}{M} \sum_{{\bf x} \in {\bf H}_{S}} \max_{{\bf y} \in {\bf H}_T} \frac{{\bf x}^{\top} {\bf y}}{||{\bf x}||~||{\bf y}||} 
\end{align}
Similarly, we calculate the other two alignment scores by:
\begin{align}
    A_{O-T} &= \frac{1}{K} \sum_{{\bf o} \in {\bf H}_{O}} \max_{{\bf y} \in {\bf H}_T} \frac{{\bf o}^{\top} {\bf y}}{||{\bf o}||~||{\bf y}||} \\
    A_{I-T} &= \max_{{\bf y} \in {\bf H}_T} \frac{{\bf h}_{I}^{\top} {\bf y}}{||{\bf h}_{I}||~||{\bf y}||}    
\end{align}
The ultimate score is a combination of the above alignments:
\begin{align}
    A = A_{S-T} + A_{I-T} + A_{O-T}   \label{eq:object}
\end{align}
The score $A$ reflects the quality of the translated captions by considering both their textual and visual information. A higher $A$ indicates that the translated caption has better quality with respect to the original caption, the relatedness between image and caption, and the similarity between image and caption at an object-level. 
Each term in $A$ measures the translation quality from a specific aspect, thereby allowing for a faithful reflection of the overall translation quality of image captions.

Practically, we followed previous work to implement the text-text alignment $A_{S-T}$ with LaBSE and implement the image-text alignment $A_{I-T}$ with CLIP. We use GRiT to implement $A_{O-T}$. GRiT will detect objects in the image and output corresponding categories. We detect the objects in the images using the GRiT model~\cite{wu2022grit} with prediction probability $>0.5$.

\subsection{Experiments}

\begin{table*}[t]
\centering
\caption{Human evaluation of the images generated by vanilla and fine-tuned diffusion models on the C$^3$ benchmark.}
\begin{tabular}{lcccccc}
\toprule
\textbf{System} & \textbf{Presence} & \textbf{Localization} & \textbf{Appropriateness} & \textbf{Aesthetics} & \textbf{Consistency} & \textbf{Cohesion} \\
\midrule
Vanilla   & 3.66 & 3.50  & 3.61 & 3.06 & 3.39 & 3.17 \\
\midrule
\multicolumn{7}{l}{\bf Fine-Tuned on Chinese-Cultural Data}\\
~~~Random     & 4.27 & 4.19  & 4.22 & 3.65 & 4.08 & 3.96  \\
~~~LaBSE      & 4.68 & 4.47  & 4.61 & 3.72 & 4.39 & 4.16  \\
~~~CLIP       & 4.66 & 4.54  & 4.56 & 3.87 & 4.38 & 4.12 \\
\midrule
~~~Ours       & \textbf{4.74}  & \textbf{4.65}  & \textbf{4.71} & \textbf{3.92} & \textbf{4.53} & \textbf{4.33} \\
\bottomrule
\end{tabular}
\label{tab:performance_on_c3}
\end{table*}

\begin{figure*}[t]
\centering
\begin{tabular}{m{0.16\textwidth}m{0.16\textwidth}m{0.16\textwidth}m{0.16\textwidth}m{0.16\textwidth}}
\midrule
\multicolumn{5}{l}{\parbox{0.9\textwidth}{A {\bf Chinese tea ceremony} with {\bf an expert} pouring {\bf tea} from {\bf a beautifully adorned teapot} into {\bf delicate cups}.}} \vspace{5pt}\\
\includegraphics[width=0.16\textwidth]{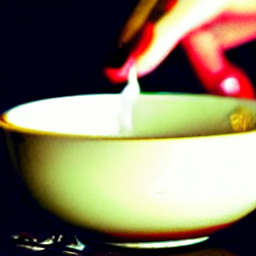} &   \includegraphics[width=0.16\textwidth]{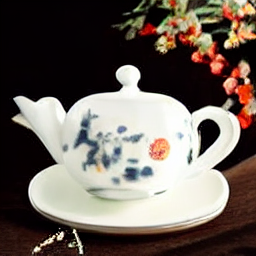} & \includegraphics[width=0.16\textwidth]{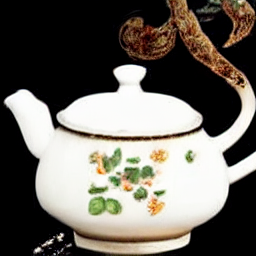} &   \includegraphics[width=0.16\textwidth]{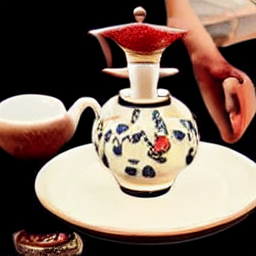} &   \includegraphics[width=0.16\textwidth]{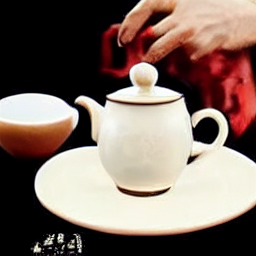}\\
\midrule
\multicolumn{5}{l}{\parbox{0.9\textwidth}{A {\bf serene Chinese garden scene}, with {\bf winding pathways},  {\bf carefully placed rocks}, and {\bf lush vegetation}, embodying the principles of harmony, balance, and connection with nature inherent in Chinese culture.}} \vspace{5pt}\\
\includegraphics[width=0.16\textwidth]{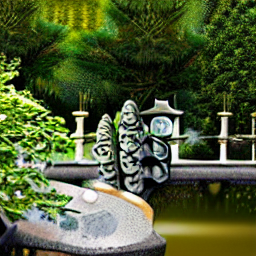} &   \includegraphics[width=0.16\textwidth]{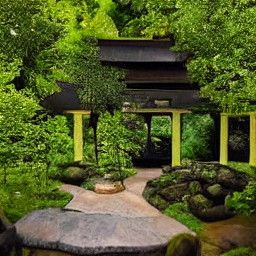} & \includegraphics[width=0.16\textwidth]{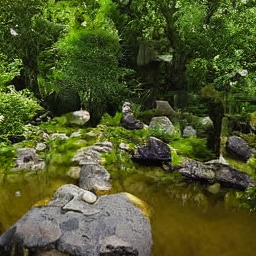} &   \includegraphics[width=0.16\textwidth]{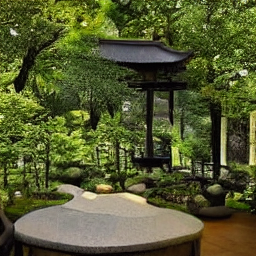} &   \includegraphics[width=0.16\textwidth]{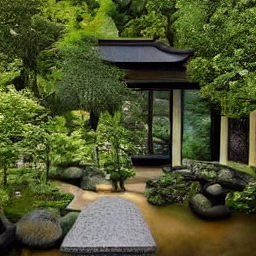}\\
\midrule
\multicolumn{1}{c}{\bf Vanilla} &   \multicolumn{1}{c}{\bf Random}  &   \multicolumn{1}{c}{\bf LaBSE}   &   \multicolumn{1}{c}{\bf CLIP}    &   \multicolumn{1}{c}{\bf Ours}\\
\end{tabular}
\caption{Example images generated by vanilla and fine-tuned diffusion models. We highlight in {\bf bold} the objects in the caption.}
\label{fig:example_generation}
\end{figure*}

In this section, we first introduce the implementation details of our experiments.
Then we present the comparison of our multi-modal metric and prior studies with human evaluation on assessing the quality of translated caption. Last, we further investigate the cross-cultural performance of different filtering metrics on the C$^3$ benchmark with quantitative results and qualitative showcases.

\paragraph{Experimental Setup}

We conduct experiments with the Stable Diffusion v1-4 model~\cite{rombach2022high}.\footnote{\url{https://github.com/CompVis/stable-diffusion}.} 
For fine-tuning the diffusion model on the Chinese cultural data, we choose the Chinese subset ({\em laion2b-zh}) of the {\em laion2b-multi} dataset\footnote{\url{https://huggingface.co/datasets/laion/laion2B-multi}.}, comprising a total of 143 million image-text pairs. We translate all image captions into English using an online translation system TranSmart~\cite{transmart}.\footnote{\url{https://transmart.qq.com}.}

We filter the full {\em laion-zh} to 300K instances with different strategies, including 1) the text-text alignment score {\bf LaBSE}~\cite{feng-etal-2022-language}; 2) the image-text alignment score {\bf CLIP}~\cite{radford2021learning}; 3) our multi-modal metric. 
We fine-tune the diffusion model on the filtered {\em laion-zh} dataset for one epoch with a batch size of 2 on 8 A100 40G GPUs. 
We use the AdamW optimizer~\cite{loshchilov2018decoupled} with a learning rate of 1e-4 for all models.

\paragraph{Assessing the Quality of Translated Caption}
We randomly sampled 500 instances from the translated {\em laion2b-zh} data, and ask human annotators to rate the quality of translated caption from two main perspectives: 1) textual translation quality, including adequacy, fluency and consistency; and 2) image correlation, including image relevance, context, and cultural appropriateness. Table~\ref{tab:caption-evaluate-guideline} in Appendix lists the evaluation guidelines.
We then scored the translated captions with different automatic metrics (e.g. LaBSE, CLIP, and Ours), and calculate their Pearson correlation with the human judgements on the above criteria.

Table~\ref{tab:direct_evaluation_translation} lists the results. Our proposed metric outperforms both LaBSE and CLIP in terms of correlation with human evaluation scores across all criteria. 
The positive correlation coefficients for our metric indicate a strong agreement between the multi-modal alignment metric and human judgments. 
This suggests that our metric is more effective in capturing the key aspects of T2I generation tasks than the other two metrics.
The results clearly demonstrate the superiority of our metric in assessing the quality of translated captions for the T2I generation tasks.

We also investigate the impact of object-text alignment score in our metric by removing it from the ultimate score (i.e. ``$-A_{O-T}$''), which is one of the key challenges in cross-cultural T2I generation. The results confirm our hypothesis: removing the object-text alignment score drastically decreases the correlation with human judgement, indicating that the alignment is essential in assessing the translated caption for cross-cultural T2I generation.

\paragraph{Performance on the C$^3$ Benchmark}

Table~\ref{tab:performance_on_c3} lists the results of different data filtering approaches on the proposed C$^3$ benchmark.
We also list the results of randomly sampling 300K instances for reference. 
Clearly, all fine-tuned models achieve significantly better performance than the vanilla model that is trained only on the English-centric data, which confirms the necessity of fine-tuning on the target cultural data for cross-cultural generation. All filtering approaches with certain metrics outperform the randomly sampling strategy, demonstrating that these metrics are reasonable for filtering low-quality instances. Our metric obtains the best results under all criteria by maintaining high-quality instances for fine-tuning.

Figure~\ref{fig:example_generation} shows some example images generated by different models. The vanilla diffusion model fails to generate Chinese-cultural elements, which can be greatly mitigated by the fine-tuned models. While CLIP and Our models successfully generate all the objects in the captions (e.g. ``tea ceremony with an expert'' and ``winding pathways, carefully placed rocks, and lush vegetation''), the elements in our images appear more natural and better-integrated. We attribute the strength of our approach to the explicit consideration of object-text alignment in data filtering.
It is also worthy noting that the proposed C$^3$ benchmark can distinguish different models by identifying model-specific weaknesses.

\section{Conclusion and Future Work}

In this work, we build a C$^3$ benchmark of challenging textual prompts to generate images in Chinese cultural style for T2I models that are generally trained on the English data of Western cultural elements. We demonstrate how the benchmark can be used to assess a T2I model's ability of cross-cultural generation from different perspectives, which reveal that the object generation is one of the key challenges. 
Based on the observation, we propose a multi-modal approach that explicitly considers object-text alignment for filtering fine-tuning data, which can significantly improves cross-cultural generation over existing metrics.
Future work include extending the C$^3$ benchmark to more non-English cultures (e.g. Arabic culture), validating our findings with more T2I models such as DALL-E 2~\cite{ramesh2022hierarchical}, and designing automatic metric and toolkit to evaluate the quality of generated images on the C$^3$ benchmark.

\bibliography{reference,custom}

\clearpage

\section{Appendix}
\label{sec:appendix}
\subsection{Human Evaluation Guidelines}
\label{app:human_eva}

Table \ref{tab:image-evaluate-guideline} provides a detailed guideline for evaluating the generated images on the C$^3$ benchmark. The evaluation is based on six criteria: Object Presence, Object Localization, Cultural Appropriateness, Visual Aesthetics, Semantic Consistency, and Cohesion. Each criterion is scored on a scale of 1 to 5, with 5 being the highest score.

\begin{itemize}
\item \textbf{Object Presence:} This criterion assesses whether all essential objects described in the caption are present and clearly recognizable in the generated image.
\item \textbf{Object Localization:} This criterion evaluates whether the spatial arrangement of objects in the image accurately represents the arrangement described in the caption.
\item \textbf{Cultural Appropriateness:} This criterion measures whether the cultural style and context described in the caption are clearly and consistently reflected in the image.
\item \textbf{Visual Aesthetics:} This criterion assesses the visual appeal and composition of the image, including color harmony, contrast, and image sharpness.
\item \textbf{Semantic Consistency:} This criterion evaluates the consistency between the image and the caption, i.e., whether all elements in the image align with and accurately represent the text.
\item \textbf{Cohesion:} This criterion measures the coherence and unity in the image, i.e., whether all elements in the image appear natural and well-integrated, creating a seamless visual scene.
\end{itemize}

Each of these criteria is crucial for evaluating the performance of T2I models, as they collectively assess the model's ability to generate images that are not only visually appealing and semantically consistent with the input text, but also culturally appropriate and coherent.

\begin{table*}[h]
\centering
\caption{Evaluation guidelines for generated images on the C$^3$ benchmark.}
\label{tab:image-evaluate-guideline}
\begin{tabular}{c p{4.5cm} p{4.5cm} p{4.5cm}}
\toprule
\textbf{Score} & \textbf{Object Presence} & \textbf{Object Localization} & \textbf{Cultural Appropriateness} \\
\midrule
5 & All essential objects are present and clearly recognizable, making the image fully consistent with the caption. & All objects are placed correctly and consistently, accurately representing the spatial arrangement described in the caption. & Cultural style and context are clearly and consistently reflected, making the image an excellent representation of the intended culture. \\
\midrule
4 & Most essential objects are present and recognizable, with only minor inconsistencies or missing details. & Most objects are placed correctly with few inconsistencies, showing a good understanding of the spatial arrangement described in the caption. & Cultural style and context are mostly well-reflected, with only minor inconsistencies or missing elements. \\
\midrule
3 & Essential objects are present, but some are missing or unclear, making the image not fully consistent with the caption. & Objects are placed reasonably well, but some inconsistencies or minor errors exist in the spatial arrangement. & Some cultural style or context is reflected, but not consistently or convincingly throughout the entire image. \\
\midrule
2 & Some essential objects are present, but not clearly recognizable or only partially visible. & Some objects are placed correctly, but most are not, showing a weak understanding of spatial arrangement. & Minimal cultural style or context is reflected, with only one or two elements hinting at the intended culture. \\
\midrule
1 & No essential objects are present in the generated image. & Objects are randomly placed with no spatial arrangement, disregarding the captions. & No cultural style or context is reflected in the generated image. \\
\bottomrule
\toprule
\textbf{Score} & \textbf{Visual Aesthetics} & \textbf{Semantic Consistency} & \textbf{Cohesion} \\
\midrule
5 & Excellent visual appeal and composition, with perfect color harmony, contrast, and image sharpness, resulting in a visually stunning image. & Complete consistency between the image and caption, with all elements aligning and accurately representing the text. & Complete coherence and unity in the image, with all elements appearing natural and well-integrated, creating a seamless visual scene. \\
\midrule
4 & Above average visual appeal and composition, with good color harmony, contrast, and image sharpness, making the image visually pleasing. & High consistency between the image and caption, with most elements aligning and only minor inconsistencies. & High coherence and unity in the image, with almost all elements appearing natural and well-integrated, creating a cohesive visual scene. \\
\midrule
3 & Average visual appeal and composition, with acceptable color harmony, contrast, and image sharpness, but lacking any outstanding qualities. & Moderate consistency between the image and caption, with some elements aligning but not enough to provide a strong connection. & Moderate coherence and unity in the image, with most objects appearing natural and well-integrated, but some inconsistencies are present. \\
\midrule
2 & Below average visual appeal and composition, with some issues in color harmony, contrast, or image sharpness. & Minimal consistency between the image and caption, with only one or two elements connecting the image to the text. & Minimal coherence or unity in the image, with some objects appearing out of place or detached from the scene. \\
\midrule
1 & Poor visual appeal and composition, with unbalanced colors, low contrast, and lack of image sharpness. & No consistency between the generated image and the caption, making the image unrelated to the text. & No coherence or unity in the image, with objects appearing disjointed and unnatural. \\
\bottomrule
\end{tabular}
\end{table*}

\subsection{Evaluation Guidelines for Translated Captions}

Table \ref{tab:caption-evaluate-guideline} provides a detailed guideline for evaluating the translated captions associated with the images. The evaluation is based on six criteria: Adequacy, Fluency, Consistency, Relevance, Context, and Cultural Appropriateness. Each criterion is scored on a scale of 1 to 5, with 5 being the highest score.

\begin{itemize}
\item \textbf{Adequacy:} This criterion assesses whether the translation accurately conveys the intended meaning of the original caption.
\item \textbf{Fluency:} This criterion evaluates the fluency of the translation, including grammar, syntax, and vocabulary.
\item \textbf{Consistency:} This criterion measures the consistency of the translations in terms of language, tone, and style.
\item \textbf{Relevance:} This criterion assesses whether the translations are relevant to the image they describe, capturing the essence of the image and all important details.
\item \textbf{Context:} This criterion evaluates whether the translations provide sufficient context for the reader to understand the image and the situation in which it was taken.
\item \textbf{Cultural Appropriateness:} This criterion measures whether the translations are appropriate for the target audience, demonstrating an understanding of the target culture and avoiding cultural references or language that could be offensive or confusing.
\end{itemize}

These criteria provide a comprehensive framework for evaluating the quality of the translated captions associated with the images, offering insights into the strengths and weaknesses of the translation process.

\begin{table*}[p]
\centering
\caption{Evaluation guidelines for the translated captions associated with the images.}
\label{tab:caption-evaluate-guideline}
\begin{tabular}{c p{4.5cm} p{4.5cm} p{4.5cm}}
\toprule
\textbf{Score} & \textbf{Adequacy} & \textbf{Fluency} & \textbf{Consistency} \\
\midrule
5 & The translation accurately conveys the intended meaning of the original caption with no errors or inaccuracies. & The translation is very well-written, with no errors in grammar, syntax, or vocabulary that could impact understanding. & The translations are consistent in language, tone, and style, with no noticeable differences.\\
\midrule
4 & The translation accurately conveys the intended meaning of the original caption with only minor errors or inaccuracies. & The translation is well-written, with only minor errors in grammar, syntax, or vocabulary that do not impact understanding. & The translations are mostly consistent in language, tone, and style, with minor differences that are hardly noticeable.\\
\midrule
3 & The translation mostly conveys the intended meaning of the original caption, but may still have some errors or inaccuracies. & The translation is generally well-written, with only a few errors in grammar, syntax, or vocabulary that do not significantly impact understanding. & The translations are generally consistent in language, tone, and style, with only a few noticeable differences.\\
\midrule
2 & The translation partially conveys the intended meaning of the original caption, but misses some important details or nuances. & The translation is somewhat fluent, but still contains some errors in grammar, syntax, or vocabulary that may make it slightly difficult to understand. & The translations are somewhat consistent, but still contain noticeable differences in language, tone, or style that may be distracting.\\
\midrule
1 & The translation does not convey the intended meaning of the original caption at all. & The translation is poorly written, with numerous errors in grammar and syntax that make it difficult to understand. & The translations are inconsistent in language, tone, or style, making them difficult to follow.\\
\bottomrule
\toprule
\textbf{Score} & \textbf{Relevance} & \textbf{Context} & \textbf{Cultural appropriateness} \\
\midrule
5 & The translations are perfectly relevant to the image they describe, capturing the essence of the image and all important details in a highly engaging way. & The translations provide perfect context for the reader to understand the image and the situation in which it was taken, leaving no room for confusion. & The translations are perfectly appropriate for the target audience, demonstrating a deep understanding of the target culture. \\
\midrule
4 & The translations are highly relevant to the image they describe, capturing the essence of the image and all important details. & The translations provide highly sufficient context for the reader to understand the image and the situation in which it was taken, with only minor room for confusion or ambiguity. & The translations are highly appropriate for the target audience, with minimal cultural references or language that could be offensive or confusing. \\
\midrule
3 & The translations are somewhat relevant to the image they describe, capturing some important details but lacking in depth or engagement. & The translations provide some context for the reader to understand the image and the situation in which it was taken, but may be somewhat confusing. & The translations are somewhat appropriate for the target audience, with some cultural references or language that may be slightly offensive or confusing. \\
\midrule
2 & The translations are minimally relevant to the image they describe, lacking important details and failing to engage the reader. & The translations provide little context for the reader to understand the image and the situation in which it was taken, leaving much room for confusion or ambiguity. & The translations are minimally appropriate for the target audience, with cultural references or language that may be offensive or confusing. \\
\midrule
1 & The translations are not relevant to the image they describe, failing to capture the essence of the image and important details. & The translations provide no context for the reader to understand the image and the situation in which it was taken, causing confusion or ambiguity. & The translations are not appropriate for the target audience, with cultural references or language that is offensive or confusing. \\
\bottomrule
\end{tabular}
\end{table*}

\end{document}